\documentclass[a4paper]{article}
\usepackage{INTERSPEECH2021}
\usepackage{cite}

\title{Advanced Long-context End-to-end Speech Recognition\\Using Context-expanded Transformers}
\name{Takaaki Hori, Niko Moritz, Chiori Hori, Jonathan Le Roux}
\address{
  Mitsubishi Electric Research Laboratories (MERL), Cambridge, MA, USA
}
\email{\{thori, moritz, chori, leroux\}@merl.com}

\begin{document}

\maketitle
\setlength{\abovedisplayskip}{4pt}
\setlength{\belowdisplayskip}{4pt}
\begin{abstract}
 This paper addresses end-to-end automatic speech recognition (ASR) for long audio recordings such as lecture and conversational speeches.
 Most end-to-end ASR models are designed to recognize independent utterances, but contextual information (e.g., speaker or topic) over multiple utterances is known to be useful for ASR.
 In our prior work, we proposed a context-expanded Transformer that accepts multiple consecutive utterances at the same time and predicts an output sequence for the last utterance, achieving 5-15\% relative error reduction from utterance-based baselines in lecture and conversational ASR benchmarks.
 Although the results have shown remarkable performance gain, there is still potential to further improve the model architecture and the decoding process. In this paper, we extend our prior work by (1) introducing the Conformer architecture to further improve the accuracy, (2) accelerating the decoding process with a novel activation recycling technique, and (3) enabling streaming decoding with triggered attention.
 We demonstrate that the extended Transformer provides state-of-the-art end-to-end ASR performance, obtaining a 17.3\% character error rate for the HKUST dataset and 12.0\%/6.3\% word error rates for the Switchboard-300 Eval2000 CallHome/Switchboard test sets. The new decoding method reduces decoding time by more than 50\% and further enables streaming ASR with limited accuracy degradation.
\end{abstract}

\section{Introduction}

Recent studies have produced different types of end-to-end models applicable for automatic speech recognition (ASR), such as connectionist temporal classification (CTC)~\cite{graves2006connectionist}, attention-based encoder decoder~\cite{bahdanau2014neural, chan2015listen}, RNN Transducer (RNN-T)~\cite{graves2013speech}, Transformer~\cite{vaswani2017attention}, and their combinations~\cite{kim2016joint_icassp2017, hori2017advances, karita2019improving,zhang2020transformer}.
Specifically, Transformer has recently provided significant performance gain over RNN-based models in major sequence-to-sequence tasks including ASR~\cite{dong2018speech, karita2019comparative}.

However, most %
ASR systems are %
designed to recognize independent utterances, despite the fact that contextual information over multiple utterances, such as information on the speaker or topic, is known to be useful for ASR.
There are several approaches to incorporating contextual information in end-to-end ASR, such as i-vector approaches that utilize speaker context~\cite{dehak2010front, audhkhasi2017direct, fan2019speaker, sari2020unsupervised} and hierarchical RNN decoders that utilize discourse context~\cite{kim2018dialog, masumura2019large}.
Besides, RNN-T and attention models have been applied to long-form ASR~\cite{chiu2019comparison}
, although the focus was
on the scalability to long-form speeches in the inference phase. 

In~\cite{hori2020transformer}, we proposed a context-expanded Transformer, which extends the Transformer model to incorporate contextual information in training and decoding to improve the recognition accuracy for lecture and conversational speeches. The proposed method concatenates multiple adjacent utterances and trains a Transformer to recognize the last of these utterances.
The previous utterances can thus be used to normalize or adapt the acoustic and linguistic features at every encoder/decoder layer for recognizing the last utterance.
Moreover, we proposed to use speaker-dependent context, i.e., concatenating the utterances spoken by the same speaker.
The proposed method achieved 5-15\% relative error reduction from utterance-based baselines in lecture and conversational ASR benchmarks.

While our approach has shown substantial accuracy improvement in long-form ASR, some issues remain to be addressed:
\begin{enumerate}
\item it is unclear whether the approach is also effective for other Transformer-based models or not, in particular whether
equivalent accuracy gains can be obtained for state-of-the-art architectures such as Conformers;
\item a high computational complexity is required in decoding due to the long speech input, as the model needs to process a long speech segment including multiple utterances to recognize the last utterance of the segment using self-attention/source-attention mechanisms, whose computational complexity increases quadratically with the segment length; 
\item the method is not yet applicable for streaming ASR, which is indispensable for online applications. %
\end{enumerate}

In this paper, we address the above mentioned issues and investigate (1) introducing the Conformer architecture~\cite{gulati2020conformer} to further improve accuracy, (2) accelerating the decoding process with a novel activation recycling technique, and (3) enabling streaming decoding with restricted self-attention and triggered attention~\cite{MoritzHR19, MoritzHR19c, moritz2020streaming}.
We demonstrate the effectiveness of the extended Transformer and its decoding method using conversational ASR benchmarks on the HKUST~\cite{liu2006hkust} and Switchboard~\cite{swbd} corpora, achieving state-of-the-art accuracy and faster decoding compared with the original approach.

\section{Context-expanded Transformer}
Transformer~\cite{vaswani2017attention} consists of encoder and decoder networks, which have deep feed-forward architectures including repeated blocks of self-attention and feed-forward layers with residual connections~\cite{he2016deep} and layer normalization~\cite{ba2016layer}.
The decoder network also features a source attention layer in each block to read the encoder's output.

\begin{figure}[t]
    \centering
    \includegraphics[width=8.0cm]{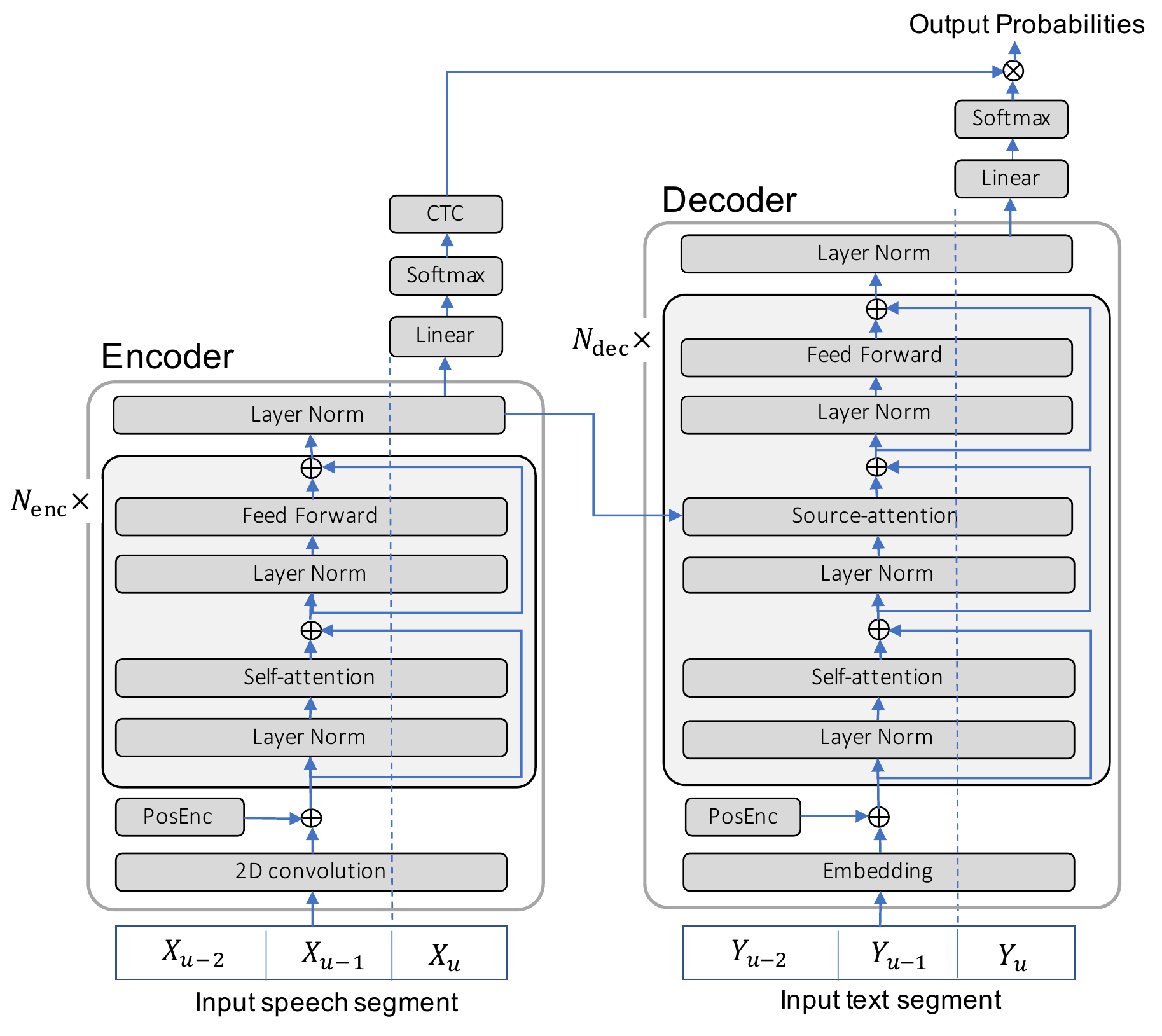}
    \caption{Context-expanded Transformer~\cite{hori2020transformer}}
    \label{fig:context-expanded-transformer}
    \vskip -5mm
\end{figure}

Figure \ref{fig:context-expanded-transformer} illustrates the context-expanded Transformer proposed in our prior work~\cite{hori2020transformer}. The network architecture is basically the same as the original Transformer, but it accepts multiple utterances at once and predicts output tokens for the last utterance using previous utterances as contextual information.

Given a feature sequence $X_u$ for a $u$-th utterance and feature sequences $X_v,\dots,X_{u-1}$ for its previous utterances, where $1\leq v < u$, we denote the input speech segment as $X_{v:u}=(X_{v},\dots,X_{u-1},X_u)$ and its corresponding output segment as $Y_{v:u}=(Y_{v},\dots,Y_{u-1},Y_u)$.
The goal of ASR here is to find the most probable token sequence $\hat{Y}_u$ for the $u$-th utterance as
\begin{align}
    \hat{Y}_u &= \mathop{\mathrm{argmax}}_{Y_u \in \mathcal{V}^*} p(Y_u|Y_{v:u-1}, X_{v:u}) \nonumber \\
              &= \mathop{\mathrm{argmax}}_{Y_u=y_{u,1:L}\in \mathcal{V}^*} \prod_{i=1}^L p(y_{u,i}| Y_{v:u-1},y_{u,1:i-1},X_{v:u}), \label{eq:seq_prob}
\end{align}
where $y_{u,1:L}$ denotes the token sequence $(y_{u,1},\dots,y_{u,L})$ of $Y_u$, and $\mathcal{V}$ is the vocabulary.

The probability of $y_{u,i}$ in Eq.~\eqref{eq:seq_prob} is computed using the Transformer.
The encoder first applies 2D convolution (Conv2D) and positional encoding (PosEnc) to all frames of $X_{v:u}$ and adds them to obtain the first hidden vector sequence
\begin{align}
    H_{v:u}^0=\mathrm{Conv2D}(X_{v:u}) + \mathrm{PosEnc}(X_{v:u}).
\end{align}
Then, it computes a hidden vector sequence in each encoder block, the sequence after the $n$-th block being obtained as
\begin{align}
    \bar{H}_{v:u}^{n-1} &= \xi(H_{v:u}^{n-1}) \\
    \tilde{H}_{v:u}^n &= H_{v:u}^{n-1}+\mathrm{MHA}(\bar{H}_{v:u}^{n-1}, \bar{H}_{v:u}^{n-1}, \bar{H}_{v:u}^{n-1})) \label{eq:enc_mha} \\
    H_{v:u}^n &= \tilde{H}_{v:u}^n+\mathrm{FFN}(\xi(\tilde{H}_{v:u}^n))),\label{eq:enc_ffn}
\end{align}
where $\mathrm{MHA}(\cdot,\cdot,\cdot)$, $\mathrm{FFN}(\cdot)$, and $\xi(\cdot)$ represent multi-head attention, feed-forward network, and layer normalization, respectively.
$\rm MHA()$ takes three arguments $Q$, $K$, and $V$, which are query, key, and value vector sequences~\cite{vaswani2017attention}. For self-attention in the encoder, these arguments are equal to $\bar{H}_{v:u}^{n-1}$.
The encoder states are obtained as the normalized output of the last block, i.e., $\bar{H}_{v:u}^{N_\mathrm{enc}}=\xi(H_{v:u}^{N_\mathrm{enc}})$, where $N_\mathrm{enc}$ denotes the number of encoder blocks.

The decoder accepts previous output sequence $(Y_{v:u-1},y_{u,1:i-1})$ and the encoder states $\bar{H}_{v:u}^{N_\mathrm{enc}}$, and estimates the probability distribution of $y_{u,i}$ in Eq.~\eqref{eq:seq_prob}.
For simplicity, we rewrite $(Y_{v:u-1},y_{u,1:i-1})$ as $y'_{u,1:k-1}$, which represents all previous tokens up to index $k-1$ in the whole segment, where $|Y_{v:u-1}| < k \leq |Y_{v:u}|$ and $k=|Y_{v:u-1}|+i$, $|Y_*|$ denoting the number of tokens in sequence $Y_*$.

The decoder first applies token embedding and positional encoding as
\begin{align}
    g^{0}_{u,1:k-1}=\mathrm{Embed}(y'_{u,1:k-1}) + \mathrm{PosEnc}(y'_{u,1:k-1}),
\end{align}
where $\mathrm{Embed}(\cdot)$ represents the token embedding.
Next, the decoder computes hidden vector $g^{n}_{u,k-1}$ in each block $n$ as 
\begin{align}
    \bar{g}^{n-1}_{u,k-1} &= \xi(g^{n-1}_{u,k-1}) \\
    \tilde{g}^{n}_{u,k-1}&=g^{n-1}_{u,k-1}+\mathrm{MHA}(\bar{g}^{n-1}_{u,k-1}, \bar{g}^{n-1}_{u,1:k-1}, \bar{g}^{n-1}_{u,1:k-1})\\
    \tilde{\tilde{g}}^{n}_{u,k-1}&=\tilde{g}^{n}_{u,k-1}+\mathrm{MHA}(\xi(\tilde{g}^{n}_{u,k-1}), \bar{H}_{v:u}^{N_\mathrm{enc}}, \bar{H}_{v:u}^{N_\mathrm{enc}}) \label{eq:src_att} \\
    g^{n}_{u,k-1}&=\tilde{\tilde{g}}^{n}_{u,k-1}+\mathrm{FFN}(\xi(\tilde{\tilde{g}}^{n}_{u,k-1})),
\end{align}
and outputs the decoder states obtained as the normalized output of the last block, i.e., $\bar{g}^{N_\mathrm{dec}}_{u,k-1}=\xi(g^{N_\mathrm{dec}}_{u,k-1})$, where $N_\mathrm{dec}$ denotes the number of decoder blocks.
Eq.~\eqref{eq:src_att} applies source attention over the encoder states, in which
$\xi(\tilde{g}^{n}_{u,k-1})$ is used for the query vector.  
Finally, we obtain the Transformer token probability distribution by applying a linear transformation and a softmax function as
\begin{align}
    p_\mathrm{trs}(y_{u,i}|&Y_{v:u-1},y_{u,1:i-1},X_{v:u}) \nonumber \\
         & =\mathrm{Softmax}(\mathrm{Linear}(\bar{g}^{N_\mathrm{dec}}_{u,|Y_{v:u-1}|+i-1})).
\end{align}

We can also utilize CTC in training and decoding similarly to the CTC-Attention approach in RNN-based architectures \cite{kim2016joint_icassp2017,hori-ACL-2017,karita2019improving}.
The CTC sequence probability can be computed as
\begin{align}
    p_\mathrm{ctc}(Y_u|X_{v:u})=\mathrm{CTC}(\mathrm{Softmax}(\mathrm{Linear}(\bar{H}_u^{N_\mathrm{enc}})), Y_u),
\end{align}
where $\mathrm{CTC}(P, Y)$ is an operation that marginalizes the posterior probabilities over all possible alignments between $P$ and $Y$ using the forward-backward algorithm~\cite{graves2006connectionist}.

For training, we use the CTC-attention loss computed as
\begin{align}
    \mathcal{L}_u=-\alpha &\log p_\mathrm{trs}(Y_u^*|Y_{v:u-1}^*,X_{v:u}) \nonumber \\
                  &- (1-\alpha)\log p_\mathrm{ctc}(Y_u^*|X_{v:u}),
                  \label{eq:loss}
\end{align}
where $Y_u^*$ and $Y_{v:u-1}^*$ are ground-truth transcripts, and $\alpha$ is a scaling factor to balance the Transformer and CTC losses. 

For decoding, we combine Transformer, CTC, and optionally LM scores to find the best hypothesis as
\begin{align}
    \hat{Y}_u=&\mathop{\mathrm{argmax}}_{Y_u \in \mathcal{V}^*}\{\lambda \log p_\mathrm{trs}(Y_u|Y_{v:u-1},X_{v:u}) \nonumber \\
    &+ (1-\lambda) \log p_\mathrm{ctc}(Y_u|X_{v:u}) + \gamma \log p_\mathrm{lm}(Y_u)\},
    \label{eq:dec_score}
\end{align}
where $\lambda$ and $\gamma$ are scaling factors to balance the model scores.
Similarly to prior studies, we employ output-synchronous beam search to efficiently find the best hypothesis \cite{hori2017advances}.
In Eq.~\eqref{eq:dec_score}, we can choose a context-dependent LM in a form of $p_\mathrm{lm}(Y_u|Y_{1:u-1})$. With an RNN-LM, contextual information beyond one utterance can be used by passing the state information from the previous utterance.
To recognize long audio recordings such as lecture and conversational speeches, we repeat the decoding process of Eq.~\eqref{eq:dec_score} in a sliding-window fashion with one-utterance shifts~\cite{hori2020transformer}.

\section{Improvements to context-expanded Transformers}
\subsection{Context-expanded Conformer}
Conformer is a variant of Transformers augmented with convolution~\cite{gulati2020conformer}, in which each encoder block has a convolution module right after the multi-head self-attention (MHSA) layer.
The convolution module consists of layer normalization, point-wise convolution, gated linear unit (GLU) activation, 1D depth-wise convolution, batch normalization, Swish activation, point-wise convolution, and dropout.
Besides, each MHSA layer in the encoder employs relative positional encoding~\cite{dai2019transformer}, which allows the self-attention module to generalize better to different input length.%
The encoder block is also extended to have a sandwich structure, where the original feed-forward layer is replaced with two half-step feed-forward layers, one before MHSA and the other after the convolution module.
It has been shown that the Conformer-based models provide substantial accuracy improvement over vanilla Transformers~\cite{gulati2020conformer,guo2020recent}.

The framework of context-expanded Transformer can be applied to the Conformer without changing its basic architecture.
The encoder steps of Eqs. \eqref{eq:enc_mha} and \eqref{eq:enc_ffn} are replaced with the sandwich structure including the convolution module as
\begin{align}
    \bar{\bar{H}}_{v:u}^{n} &= \bar{H}_{v:u}^{n-1} + \frac{1}{2}\mathrm{FFN}(\bar{H}_{v:u}^{n-1})\\
    \tilde{H}_{v:u}^{n} &= \bar{\bar{H}}_{v:u}^{n}+\mathrm{MHA}(\xi(\bar{\bar{H}}_{v:u}^{n}), \xi(\bar{\bar{H}}_{v:u}^{n}), \xi(\bar{\bar{H}}_{v:u}^{n})) \\
    \tilde{\tilde{H}}_{v:u}^{n} &= \tilde{H}_{v:u}^{n} + \mathrm{Conv}(\xi(\tilde{H}_{v:u}^{n})) \\
    H_{v:u}^n &= \tilde{\tilde{H}}_{v:u}^{n}+\frac{1}{2}\mathrm{FFN}(\xi(\tilde{\tilde{H}}_{v:u}^{n})),
\end{align}
where $\mathrm{Conv}(\cdot)$ denotes the convolution module. 

\subsection{Efficient decoding with activation recycling}
Computational complexity of multi-head attention increases quadratically with the sequence length.
With context expansion, the input sequence is approximately 4 times longer than single utterances in a standard setting of context-expanded Transformer,
obviously causing a major increase in decoding time.

Considering a sliding-window decoding with one-utterance shift moving focus to a new utterance $X_{u}$, instead of computing new hidden activation vectors for previous utterances $X_{v:u-1}$ taking into account the new current utterance $X_{u}$ as in \cite{hori2020transformer}, we can reuse the hidden activation vectors computed for each previous utterance when it was decoded. 
Accordingly, the hidden activations in the encoder block can be computed as
\begin{align}
    \tilde{H}_u^{n}&=H_u^{n-1}+\mathrm{MHA}(\bar{H}_u^{n-1}, \bar{H}_{v:u}^{n-1}, \bar{H}_{v:u}^{n-1}) \\
    H_u^n&=\tilde{H}_u^{n}+\mathrm{FFN}(\xi(\tilde{H}_u^{n})) \label{eq:enc_recycle}\\
    H_{v:u}^{n}&=({\cal H}_{v:u-1}^n, H_u^n),
\end{align}
where hidden activations ${\cal H}_{v:u-1}^n$ have already been computed and cached in the memory when the previous utterances were decoded. The activations thus need to be computed only for the current utterance $X_u$. Furthermore, in multi-head attention, the length of query is reduced from the segment length $|H_{v:u}^n|$ to the length of the current utterance $|H_u^n|$, i.e., the computational complexity is reduced from $O(|H_{v:u}^n|^2)$ to $O(|H_u^n|\times|H_{v:u}^n|)$.

The computation for other layers including feed-forward layers reduces to the same level as ordinary utterance-based ASR.
For recycling the cached activations, it is important to employ relative positional encoding to make the activations independent of their positions. Moreover, as cached activations cannot have any information from future utterances, we need to omit backward self-attention across utterances by masking such connections during training.

This recycling mechanism can also be used for the decoder blocks, where relative positional encoding is also needed but the backward self-attention does not have to be considered since the decoder uses only forward self-attention.
Furthermore, we can reduce the computation for source attention by limiting the attention span to only the current utterance, i.e., the computational complexity of source attention can be reduced from $O(|Y_{v:u}|\times |H_{v:u}^{N_{enc}}|)$ to $O(|Y_u|\times|H_u^{N_{enc}}|)$, which is the same as that of the utterance-based ASR.
The same recycling mechanism can be applied for context-expanded Conformers as well.

\subsection{Streaming decoding with triggered attention}
Streaming decoding is regarded as an essential function in ASR systems.
However, most end-to-end models are designed to access the full input sequence even for predicting the first token. This means that ASR decoding cannot start until the utterance end point is detected, leading to a substantial delay in the ASR output especially for long utterances.  
Recent studies have proposed efficient streaming algorithms for end-to-end ASR \cite{MoritzHR19, MoritzHR19c, moritz2020streaming, he2019streaming, narayanan2019recognizing, tsunoo2021streaming}, which enable  processing of the input speech in a streaming fashion and reduce the latency.
For joint CTC-attention-based models, we have proposed the triggered attention technique~\cite{MoritzHR19, MoritzHR19c, moritz2020streaming}, which utilizes CTC during training and inference to estimate a token emission timing and activate the attention decoder accordingly. Triggered attention provides time-synchronous decoding using CTC prefix beam search extended with on-the-fly attention decoder rescoring.

In this paper, we introduce triggered attention into context-expanded Transformers to realize high-accuracy streaming ASR. 
The training is performed on concatenated utterances and enforces time restriction on the self- and source-attention layers by masking attention weights to simulate a situation where future context is not available while still considering several look-ahead frames.
Activation recycling is also employed in the decoding process.

\subsection{Related work}
Apart from our own prior work~\cite{hori2020transformer}, prior studies on end-to-end ASR for long-form speeches~\cite{kim2018dialog,masumura2019large,chiu2019comparison,narayanan2019recognizing} rely on attention-based encoder-decoders or RNN-T models, while we here explore the use of
Transformer-based encoder-decoders and their extensions.

Regarding the activation recycling technique, a similar method has already been introduced in the Transformer-XL language model (LM)~\cite{dai2019transformer}. However, our Transformer has an encoder-decoder architecture and reuses activations on an utterance basis rather than fixed-sized text blocks as in Transformer-XL, since utterances are suitable as a basic processing unit for ASR. 
Moreover, we forbid backward self-attention across utterances in the encoder and apply within-utterance source attention in the decoder during training, both of which have not been considered in Transformer-XL.

Streaming decoding is an active area of research in end-to-end ASR 
\cite{he2019streaming, narayanan2019recognizing, tsunoo2021streaming}. 
One prior work did explore a streaming technique for long-form ASR with RNN-T~\cite{narayanan2019recognizing}, in which the decoding method depends on the RNN architecture, where the encoder and decoder receive state information from the previous utterance, but that technique is not applicable to Transformers.

\section{Experiments}
\subsection{Experimental setup}
We conducted several experiments using conversational ASR benchmarks on the HKUST~\cite{liu2006hkust} and Switchboard~\cite{swbd} corpora, which consist of 200 hours and 300 hours of 8 kHz telephone conversations in Mandarin Chinese and English, respectively.

The Kaldi toolkit~\cite{Povey_ASRU2011} was used to extract 80-dimensional log mel-filter bank acoustic features plus three-dimensional pitch features.
We trained Transformers with the architecture in ESPnet \cite{watanabe2018espnet,karita2019comparative}.
The encoder had one Conv2D module followed by 12 encoder blocks ($N_\mathrm{enc}=12$).
The Conv2D included a 2-layer 2D-CNN with 256 channels, a kernel size of $3 \times 3$, a stride of size $2$, and ReLU activation, which outputs a 256-dimensional vector sequence with the utterance length reduced by a factor of 4.
We employed multi-head attention with 4 heads of 256 dimensions. The feed-forward network had one hidden layer with 2,048 units and ReLU non-linearity. 
The decoder had a token embedding layer followed by 6 decoder blocks ($N_\mathrm{dec}=6$).
The self-attention, source attention, and feed forward layers in the decoder had the same dimensions as those in the encoder.
The output dimension was dependent on the number of unique tokens in the task,with 3,653 characters in HKUST and 1,996 word pieces in Switchboard. 

We basically followed the default configuration of ESPnet recipes~\cite{watanabe2018espnet}, where speed perturbation and SpecAugment~\cite{park2019specaugment} were applied for both data sets. 
Baseline Transformers were trained with independent utterances without context. 
To train Transformers with the proposed method, we expanded each utterance to a 20-second segment by concatenating it with previous utterances.
Unlike our prior work~\cite{hori2020transformer}, we trained the context-expanded models with CTC-attention loss in Eq.~\eqref{eq:loss} by fine-tuning a corresponding utterance-based model using Adam optimizer, since our preliminary experiments indicated that fine-tuning is more stable and faster than training from scratch as we have done before.
Finally, we averaged the top 10 models based on validation accuracy for recognition.

We also trained RNN-LMs using transcripts for HKUST and Switchboard, further adding transcripts from the Fisher corpus for Switchboard. The LMs had 2 LSTM layers with 650 cells for HKUST and 1,024 cells for Switchboard.
The transcripts were concatenated in the same manner as in context-expanded Transformer training.
ASR performance was measured by character error rate (CER) or word error rate (WER).

\subsection{Results}

\begin{table}[t]
\centering
\caption{Recognition error rate vs.\ decoding time in HKUST and Switchboard benchmarks. The decoding time is represented in real-time factor (RTF), when decoded with a single-thread process on an Intel Core i7-3970X CPU @ 3.50GHz.}
\label{table:error-rate-and-decoding-speed}
\vskip -3mm
\resizebox{.99\linewidth}{!}{
\begin{tabular}{lcccc}
\toprule
        & \multicolumn{2}{c}{HKUST}& \multicolumn{2}{c}{Switchboard} \\
         \cmidrule(lr){2-3}\cmidrule(lr){4-5}
        & \multicolumn{2}{c}{dev}  & \multicolumn{2}{c}{CH / SWB} \\
         \cmidrule(lr){2-3}\cmidrule(lr){4-5}
        & CER [\%] & RTF & WER [\%] & RTF  \\
\midrule
Baseline Transformer    &  21.2  &  0.44 & 15.5 / 7.8 & 0.66 \\
Context-Ex.\ Transformer &  18.9  &  1.26 &  14.0 / 7.1 & 1.89 \\
+ Activation recycling  &  {\bf 18.8}  & {\bf 0.61} & {\bf 14.0} / 7.2 & {\bf 0.93} \\
\midrule
Baseline Conformer      &  20.0  & 0.46  & 13.9 / 6.8 & 0.68 \\
Context-Ex.\ Conformer   & 17.3 & 1.31   & 12.1 / 6.3 & 2.05 \\
+ Activation recycling  & {\bf 17.3} & {\bf 0.62} & {\bf 12.0 / 6.3} & {\bf 1.02} \\
\midrule
ESPnet Conformer \cite{guo2020recent} & 22.2 & - & 15.0 / 7.1 & - \\
Att. enc-dec.~\cite{park2019specaugment} &   -  & - & 14.0 / 6.8 & - \\
Att. enc-dec.~\cite{tuske2020single} & - & - & 12.5 / 6.4 & - \\
\bottomrule
\end{tabular}
}
\vskip -6mm
\end{table}
Table \ref{table:error-rate-and-decoding-speed} shows recognition error rate and decoding speed for utterance-based (Baseline) and context-expanded (Context-Ex.) models.
For both datasets, context expansion provides substantial relative error reduction ranging from 5\% to 13.5\% for both Transformers and Conformers.
The context-expanded Conformer achieved 17.3\% CER on the HKUST dev set and 12.0\% / 6.3\% WER on the CallHome (CH) and Switchboard (SWB) subsets of the Switchboard Eval2000 test set.
These numbers are better than those reported in other papers \cite{guo2020recent, park2019specaugment, tuske2020single}.
We also evaluated the effect of activation recycling in beam search decoding \cite{seki2019vectorized} with beam size 10. In decoding, we increased the segment size to 25 seconds for context-expanded models on Switchboard, since this provides slightly better WERs for the validation and test sets.
Without recycling, the RTF increases roughly by a factor of 3 from the baseline.
With recycling, however, the decoding time can be reduced to half of the original time with almost no increase of errors. The recycling mechanism thus effectively works for context-expanded Transformers. Consequently, we can perform context-expanded ASR with a 35-50\% increase of computation time from the baseline on a single CPU core.

\begin{table}[t]
\centering
\caption{Streaming ASR results. Numbers indicate CERs [\%] for HKUST and WERs [\%] for Switchboard. ``\%inc'' indicates the error increase ratio from the full-sequence model.}
\label{table:streaming-asr}
\vskip -3mm
\resizebox{.99\linewidth}{!}{
\setlength{\tabcolsep}{4pt}
\begin{tabular}{llll}
\toprule
                        & \multicolumn{1}{c}{HKUST} & \multicolumn{2}{c}{Switchboard} \\
                        \cmidrule(lr){2-2}\cmidrule(lr){3-4}
                        &  dev (\%inc) & CH (\%inc) & SWB (\%inc) \\
\midrule
Baseline Transformer     & 21.2       & 15.5        & 7.8      \\
+ Triggered attention    & 23.1~(9.0) & 17.9~(15.5) & 9.0~(15.4) \\
Context-Ex. Transformer  & 18.8       & 14.0        & 7.2      \\
+ Triggered attention    & {\bf 20.2~(7.4)} & {\bf 15.5~(10.7)} &  {\bf 8.4}~(16.7) \\
\midrule
Baseline Conformer       & 20.0        & 13.9        & 6.8 \\
+ Triggered attention    & 22.6~(13.0) & 16.8~(20.0) & 8.0~(17.6) \\
Context-Ex. Conformer    & 17.3        & 12.0        & 6.3 \\
+ Triggered attention    & {\bf 19.3~(11.6)} & {\bf 14.2~(18.3)} & {\bf 7.2~(14.3)} \\
\bottomrule
\end{tabular}
}
\vskip -6mm
\end{table}
Next, we investigate streaming ASR with triggered attention.
In triggered attention decoding, we applied a 1-frame self-attention look-ahead at each encoder layer, which results in a 12-frame delay for the 12-layer encoder.
We also used a 12-frame look-ahead on the encoder frames in source attention by the decoder.
Accordingly, this configuration requires a 480ms + 480ms theoretical delay, since one encoder frame corresponds to 40ms. For Conformers, we restricted the depth-wise convolution to use only past frames. The error rates are summarized in Table \ref{table:streaming-asr}.
The results show that the context-expanded models can reduce the errors in streaming ASR as well.
We also observe a slight deterioration of error rate from full-sequence (i.e., not streaming) ASR, as is usually observed due to
a lack of future information in streaming ASR.
But, according to the error increase ratio (\%inc), the context-expanded models tend to mitigate the increase in errors compared to the baseline models. 
This fact suggests that the context-expanded Transformers are more effective in streaming ASR, since the utterance-based models can use only little information at an early stage of decoding while the context-expanded models can utilize more contextual information from previous utterances.

\section{Conclusions}
In this paper, we have extended our prior work on context-expanded Transformers, which exploit contextual information of previous utterances to improve ASR accuracy for the current utterance.
We have investigated three extensions: (1) Conformer architecture for further accuracy improvement, (2) accelerated decoding by activation recycling, and (3) streaming decoding with triggered attention.
We have demonstrated that the extended Transformers provide state-of-the-art end-to-end ASR performance, achieving 17.3\% CER on the HKUST dev set and 12.0\% / 6.3\% WER on the Switchboard-300 Eval2000 CallHome/Switchboard test sets. The new decoding method reduced decoding time to under 50\% of that of the original method and further enabled streaming ASR without considerable accuracy degradation.
  
\bibliographystyle{IEEEtran}

\bibliography{mybib}

\end{document}